\documentclass[letterpaper, 10 pt, conference]{ieeeconf}  

\IEEEoverridecommandlockouts                              

\usepackage{multicol}
\usepackage[bookmarks=true]{hyperref}
\usepackage{color}
\usepackage{graphicx}
\usepackage{bm}
\usepackage{tabularx}
\usepackage{floatrow}
\usepackage[utf8]{inputenc}
\usepackage{amsmath}
\usepackage[utf8]{inputenc} 
\usepackage[T1]{fontenc}    
\usepackage[mathscr]{eucal}
\usepackage{subfig}
\usepackage{caption}
\usepackage{amsfonts}
\usepackage{dblfloatfix} 
\usepackage{epsfig}
\usepackage{makecell}
\usepackage{multirow}
\usepackage{hyperref}
\usepackage{dsfont}
\usepackage{breqn}
\usepackage{mathtools,algorithm}
\usepackage[noend]{algpseudocode}
\title{\LARGE \bf
Learning Safe Unlabeled Multi-Robot Planning with Motion Constraints}

\author{Arbaaz Khan$^{1}$, Chi Zhang$^{1}$, Shuo Li$^{1}$, Jiayue Wu$^{1}$, Brent Schlotfeldt$^{1}$, \\
Sarah Y. Tang$^{2}$, Alejandro Ribeiro$^{1}$, Osbert Bastani$^{1}$, Vijay Kumar$^{1}$ 
\thanks{$^{1}$Authors are with GRASP Lab, University of Pennsylvania, USA
        {\tt\small arbaazk@seas.upenn.edu}}%
\thanks{$^{2}$Sarah Y. Tang is with Nuro, Palo Alto, CA, USA}%
\thanks {This paper has been accepted to the IEEE/RSJ International Conference on Intelligent Robots and Systems 2019.}
}

\begin{document}

\maketitle
\thispagestyle{empty}
\pagestyle{empty}

\begin{abstract}
In this paper, we present a learning approach to goal assignment and trajectory planning for unlabeled robots operating in 2D, obstacle-filled workspaces. More specifically, we tackle the unlabeled multi-robot motion planning problem with motion constraints as a multi-agent reinforcement learning problem with some sparse global reward. In contrast with previous works, which formulate an entirely new hand-crafted optimization cost or trajectory generation algorithm for a different robot dynamic model, our framework is a general approach that is applicable to arbitrary robot models. Further, by using the velocity obstacle, we devise a smooth projection that guarantees collision free trajectories for all robots with respect to their neighbors and obstacles. The efficacy of our algorithm is demonstrated through varied simulations. A video describing our method and results can be found \href{https://www.youtube.com/watch?v=ggTuBdAbrIU}{here.}
\end{abstract}


\section{Introduction}
In many applications in robotics such as formation flying~\cite{desai2001modeling,alonso2015multi} or perimeter defense and surveillance~\cite{saldana2016dynamic}, there exist teams of interchangeable robots operating in complex environments. In these scenarios, the goal is to have a team of robots execute a set of identical tasks such that each robot executes only one task, but it does not matter which robot executes which task. One example of such a problem is the concurrent goal assignment and trajectory planning problem where robots must simultaneously assign goals and plan motion primitives to reach assigned goals. 

Solutions to this unlabeled multi-robot planning problem must solve both the goal assignment and trajectory optimization problems. It has been shown that the flexibility to freely assign goals to robots allows for polynomial-time solutions under certain conditions~\cite{adler2015efficient, macalpine2015scram,yu2012distance}. Nonetheless, there are still significant drawbacks to existing approaches. Solutions with polynomial-time complexities depend on minimum separations between start positions, goal positions, or robots and obstacles~\cite{adler2015efficient,macalpine2015scram,turpin2014capt}. Motion plans generated by graph-based approaches for this problem such as those proposed in~\cite{yu2012distance,yu2013multi} are limited to simple real-world robots that have approximately first-order dynamics. Closest to our work,~\cite{turpin2014goal} proposes an algorithm to coordinate unlabeled robots with arbitrary dynamics in obstacle-filled environments. However, it assumes the existence of a single-robot trajectory optimizer that can produce a candidate trajectory for a robot to any given goal, which is in itself a difficult research problem. Furthermore, the algorithm is a priority-based method that depends on offsetting the times at which robots start traversing their trajectories to guarantee collision avoidance. In the worst case, it degenerates into a completely sequential algorithm where only one robot is moving at a time. 
\begin{figure}[t!]
  \centering
  \includegraphics[scale=0.41]{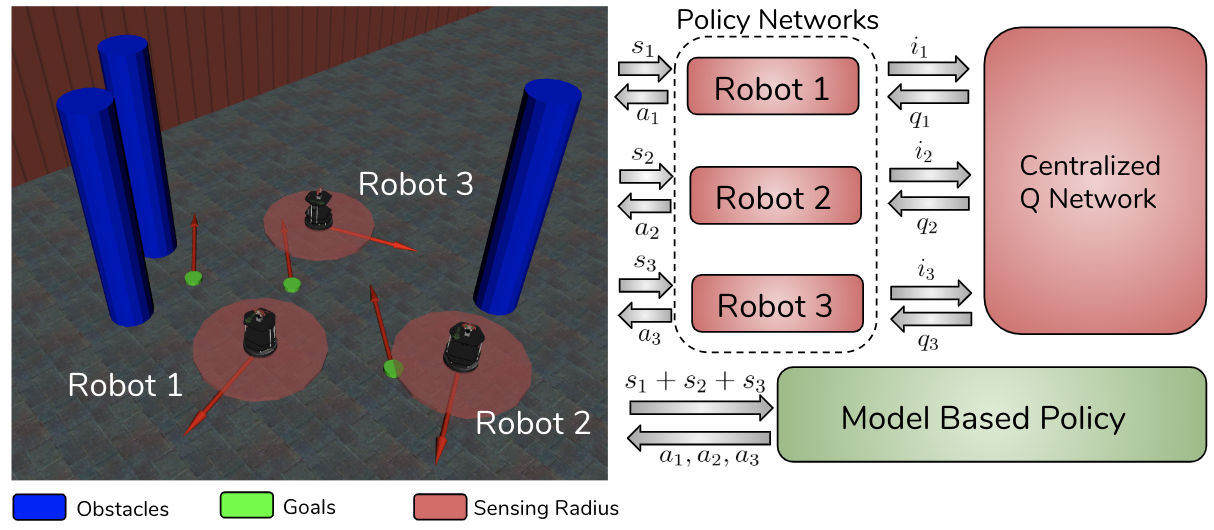}
  \caption{\textbf{Learning Unlabeled Motion Planning} Robots observe their own state and velocity, relative positions of all goals and other entities within a sensing region. For robot $n$, this information is compiled into a vector $s_{n}$. Each robot uses its own policy network to compute an action $a_n$. During training, policy networks also exchange information $i$ with a centralized Q-network which uses information from all robots to compute robot specific Q functions $q_n$. These Q functions are used by the individual robot functions to update their policies. To guarantee safety, a model based policy runs in the background. \label{fig:page1}}
 \vspace{-0.1cm}
\end{figure}

In light of these works, we propose a novel learning-based framework for goal assignment and trajectory optimization for a team of identical robots with arbitrary dynamics operating in obstacle-filled work spaces. Firstly, it is observed that the unlabeled multi-robot planning problem can be recast as a multi-agent reinforcement learning (MARL) problem. Robots are given their own state, configuration of obstacles and information about other robots within some sensing radius and configuration of all goals in the environment. The objective then is to learn policies that couple the assignment and trajectory generation for each robot. It is important to note that in such an approach, since we do not have goals assigned beforehand, it is non-trivial to assign individual rewards to each robot. Instead, we simply assign a global reward to all robots that takes into account if all robots have reached goals in a collision free manner. This global reward then forces each robot to learn policies such that all robots can reach a goal without colliding or having to communicate with each other. Thus, by casting the concurrent goal assignment problem as a MARL problem, we attempt to learn policies that maximize this global reward in minimal time. A centralized training, decentralized execution strategy is used to train policies for each robot. This builds on a body of work in MARL that employ such a centralized training, decentralized execution strategy~\cite{maddpg,foerster2017counterfactual}. 

When utilizing a deep reinforcement learning (RL) model, one loses optimality guarantees as well as any guarantees for collision free trajectories. To ensure one still has collision free trajectories, we make use of an analytical model based policy that runs in the background and checks if the target velocities produced by the robot are "safe". For each robot, velocity obstacles are computed at every instant in time. The velocity obstacle divides the set of all possible velocities into safe and unsafe sets. If the velocity computed by the learned model lies in the unsafe set, it is projected back into the safe set. We also ensure that this projection preserves certain properties such as smoothness of the transition function in order to be compatible with the learning process. Thus, when using this model based policy in conjunction with the policy learned, we are guaranteed almost safe trajectories at every instant and it is empirically shown that the computed policies converge to the desired values. 

Thus, the main contributions of our algorithm are : 1) Capability to extend to arbitrary robot dynamics. We test with both holonomic and non-holonomic robot models 2) There is no need to re-train if the obstacles are moved around or if the robot's start and goal positions are changed. This is in contrast to any model-based approach where one would need to recompute the assignment and regenerate the trajectory.  3) It  retains the safety guarantees of model-based methods, 4) We show experimentally that better performance based on total time to reach goals is achieved than model based methods alone. 


\section{Problem Formulation} \label{sec:Preliminaries}
Consider a two dimensional Euclidean space with $N$ homogeneous disk-shaped \color{black} robots of radius $R$ indexed by $n$ and $M$ goal locations indexed by $m$. Goal location $m$ is represented as a vector of its position ($x_m,y_m$) and heading ($\theta_m$) in the plane. Thus, goal $m$ is :
\begin{equation}
    \textbf{g}_m = [x_m,y_m,\theta_m]
\end{equation}
The full goal state vector, \textbf{G} $\in \mathbb{SE}(2)^M$ is given as :
\begin{equation}
    \textbf{G}=[\textbf{g}_1,\textbf{g}_2,\ldots,\textbf{g}_M]
\end{equation}
Let robot $n$ be equipped with a sensor with range $R_s$ ($R_s > R$). The data from the sensor is denoted as $I_n$. We assume each robot has full information about the location of the goals and full information about its own pose. Thus, the observation of the $n$th robot at time $t$ is then given as :
\begin{equation}
\label{eq:state}
    \textbf{o}_t^{n} = [\textbf{p}_n(t),v_{n}(t),\omega_n(t), I_n,\textbf{G}]
\end{equation}
where $\textbf{p}_n(t) \in \mathbb{SE}(2)$ is a vector of the position ($x_n(t),y_n(t)$) and heading ($\theta_n(t)$) of robot  $n$ in the plane and is given as :
\begin{equation}
    \textbf{p}_n(t) = [x_n(t),y_n(t),\theta_n(t)]
\end{equation}
The linear velocity of the robot at time $t$ is denoted as $v_{n}(t)$ and the angular velocity is denoted as $\omega_n(t)$. 
Let $\mathcal{S}$ be the set of states describing all possible configurations of all robots. 
For each robot, given an action $a_t^n \in \mathcal{A}_n$ where $\mathcal{A}_n$ describes all possible actions for robot $n$, the state of the robot evolves according to some stationary dynamics distribution with conditional density $p(\textbf{o}_{t+1}|\textbf{o}_t,a_t)$. These dynamics can be linear or non-linear. Further let $\mathcal{A} =[\mathcal{A}_1,\mathcal{A}_2,\ldots\mathcal{A}_n]$ be the set of all possible actions of all robot. We define some arbitrary scalar $\delta > 0$ such that the necessary and sufficient condition to ensure collision avoidance is given as :
\begin{equation}
\label{eq:collision}
\begin{split}
     E_c(p_i(t),p_j(t)) > 2R+\delta, \\ \forall i \neq j \in \{1,\ldots N\}, \forall {t}
\end{split}
\end{equation}
\color{black}
where $E_c$ is the euclidean distance. Lastly, we define the assignment matrix $\phi(t) \in \mathbb{R}^{N \times M}$ as 
\begin{equation}
    \phi_{ij}(t) = 
    \begin{cases}
    1, &\text{if }  E_c(p_i(t),g_j) + D_c(p_i(t),g_j) \leq \epsilon \\
    0, &\text{otherwise}
    \end{cases}
\end{equation}
where $D_C$ is the cosine distance and $\epsilon$ is some threshold. In this work, we consider the case when $N=M$. The necessary and sufficient condition for all goals to be covered by robots at some time $t=T$ is then:
\begin{equation}
\label{eq:stopping}
    \phi(T)^\top\phi(T) = \textbf{I}_{N}
\end{equation}
where $\textbf{I}$ is the identity matrix. 
Thus, the problem statement considered in this paper can be defined as : \\ 

\textbf{Problem 1.} Given an initial set of observations $\{\textbf{o}_{t_0}^1,\ldots, \textbf{o}_{t_0}^N\} $ and a set of goals $\textbf{G}$, compute a set of functions $\mu_n(\textbf{o}_t^n)$  $\forall n={1,\ldots,N}, 
\forall t$ such that applying the actions $\{a_1^t,\ldots,a_N^t\} :=\{\mu_1(\textbf{o}_t^1),\ldots \mu_N(\textbf{o}_t^N)\}$ results in a sequence of observations each satisfying Eqn. \ref{eq:collision} and at final time $t=T$ satisfies Eqn. \ref{eq:stopping}. 
In the next section we outline our methodology to easily convert this problem to a Markov game for MARL and outline our methodology to compute policies such that the constraints in Eqn. \ref{eq:collision} and Eqn. \ref{eq:stopping} are satisfied.
\vspace{0.15cm}
\section{Learning Unlabeled Multi-Robot Planning}\label{sec:Algo}
\subsection{Markov Games for Multi-Robot Planning}
One can reformulate the unlabeled multi-robot planning problem describe in \textbf{Problem 1} as a Markov game~\cite{littman1994markov}. A Markov game for $N$ robots is defined by a tuple $\{\mathcal{S},\mathcal{A},\mathcal{T},\mathcal{O}, \mathcal{R},\gamma\}$. $\mathcal{S}$ describes the full true state of the environment. We observe corresponds to the full state space in the problem setup.  Similarly we observe that the action space $\mathcal{A}$ is the same as defined before in \textbf{Problem 1}. The transition function $\mathcal{T}:\mathcal{S}\times\mathcal{A}\rightarrow\mathcal{S}$ is equivalent to the dynamics distribution $p(\textbf{o}_{t+1}|\textbf{o}_t,a_t)$ describe in Section \ref{sec:Preliminaries}. In the Markov game, the environment is partially observed. At every timestep, each robot can only observe some small part of the environment. This is also defined in our problem and for every timestep $t$ we can simply set the observation for robot $\mathcal{O}_n=\textbf{o}_t^n$, and $\mathcal{O} = \{\mathcal{O}_1,\ldots,\mathcal{O}_n\}$. $\gamma$ is a discount factor and can be set close to one. In the Markov game, each robot $n$ obtains a reward function as a function of robots state and its action. We propose formulating a reward structure that satisfies the constraints in Eqn. \ref{eq:collision} and Eqn. \ref{eq:stopping}. 
\begin{equation}
\label{eq:maxreward}
    r(t) =
    \begin{cases}
    \alpha &\text{if }  \phi(t)^\top\phi(t) = \textbf{I}_N\\
   -\beta, &\text{if }  \text{any collisions}\\
   0  &\text{otherwise}
    \end{cases}
\end{equation}
where $\alpha$ and $\beta$ are some positive constants. It is important to note that this reward structure is global and is given to all robots \{$r(t) = r_1(t),=,r_n(t)$\}. By using a global reward we remove the need for any carefully designed heuristic function. In the Markov game, the solution for each robot $n$ is a policy $\pi_n(a^n|\textbf{o}^n)$ that maximizes the discounted expected reward $R_n = \sum_{t=0}^T\gamma^t r_n(t)$. Once again, we draw parallels between the Markov game and \textbf{Problem 1} and set $\mu_n = \pi_n$. Thus, we can conclude the solution of the Markov game for Multi-Robot Planning is the solution for the Unlabeled multi-robot planning considered in \textbf{Problem 1}. 
\subsection{Learning Policies for Continuous Actions}
Consider a single robot setting. The MDP assosciated with a single robot is given
as $\mathcal{M}_t(\mathcal{O},\mathcal{A},\mathcal{T}_1,r,\gamma)$ ($\mathcal{T}_1$ is the transition function associated with just the robot under consideration). 
The goal of any RL algorithm is to find a stochastic policy $\pi(\textbf{o}_t|a_t;\theta)$ (where $\theta$ are the parameters of the policy) that maximizes the expected sum of rewards :
\begin{equation}
\label{eq:rl}
\max_{\theta}  \mathbb{E}_{\pi(a_t|\textbf{o}_t;\theta)}[\sum_{t} r_t]  \enspace
\end{equation}
Policy gradient methods look to maximize the reward by estimating the gradient and using it in a stochastic gradient ascent algorithm. A general form for the policy gradient can be given as:
\begin{equation}
\hat{g} = {\mathbb{E}}_t[\nabla_{\theta}\text{log} \pi_{\theta}(a_t|\textbf{o}_t){Q}_t ^{\pi}(\textbf{o}_t,a_t)]
\end{equation}
where $Q^{\pi}(o_t,a_t)$ represents the action value function (estimate of how good it is to take an action in a state) 
\begin{equation}
Q^{\pi}(\textbf{o}_t,a_t) := \mathbb{E}_{\textbf{o}_{t+1:\infty},a_{t+1:\infty}} \Big[ \sum_{l=0}^{\infty} r_{t+l} \Big] 
\end{equation}
Bellman equations also give us the recursive rule for updating the action value function given as :
\begin{dmath}\label{eq:bellman}
    Q^{\pi}(\textbf{o}_t,a_t) = \mathbb{E}_{\textbf{o}_{t+1:\infty},a_{t:\infty}} \Big[ r(\textbf{o}_t,a_t) +\\ \gamma\mathbb{E}_{{\textbf{o}_{t+1:\infty},a_{t+1:\infty}}}(Q^{\pi}(\textbf{o}_{t+1},a_{t+1})) \Big] 
\end{dmath}
where $r(\textbf{o}_t,a_t)$ is the reward for executing action $a_t$ in $\textbf{o}_t$. 
The gradient $\hat{g}$ is estimated by differentiating the objective wrt $\theta$: 
\begin{equation}
L^{PG}(\theta)=\mathbb{E}[\text{log}(\pi_{\theta}(a_t|\textbf{o}_t)Q^{\pi}(\textbf{o}_t,a_t)]
\end{equation}
In order to extend this to continuous actions and continuous deterministic policies,~\cite{lillicrap2015continuous} propose the the Deep Deterministic Policy Gradient (DDPG) algorithm for continuous actions 
and deterministic policies. The algorithm maintains an actor function (parameterized by $\theta^{\pi}$) that estimates the deterministic continuous policy $\pi$. In addition, it also maintains a critic function (parameterized by $\theta^{Q}$) that estimates the action value function. The critic function is updated by using the Bellman loss as in Q-learning~\cite{sutton1998reinforcement} (Eqn. ~\ref{eq:bellman}) and the actor function is updated by computing the following  policy gradient :
\begin{equation}
\label{eq:ddpggradient}
    \hat{g} = {\mathbb{E}}_t[\nabla_{\theta}\text{log} \pi_{\theta}(a_t|\textbf{o}_t)\nabla_a {Q}_t ^{\pi}(\textbf{o}_t,a_t)]
\end{equation}
The DDPG algorithm is an off-policy algorithm and samples trajectories from a replay buffer of experiences stored in a replay buffer. Similar to DQN~\cite{dqn} it also uses a target network to stabilize training. 

A natural question to ask at this point is, why not treat every robot in the space as an entity operating independently and learn this DDPG algorithm for each robot and in fact this exact idea has been proposed in \textit{Independent Q-Learning}~\cite{tan1993multi}. However, there are two major drawbacks to this approach. When operating in high dimensional continuous spaces with sparse rewards, the lack of information sharing between the robots makes it difficult to learn any co-ordination between robots. Further, as each robot's policy changes during training, the environment becomes non-stationary from the perspective of any individual robot (in a way that cannot be explained by changes in the robots’s own policy). This is the non-stationarity problem in multi-agent learning.  

\subsection{Learning Continuous Policies for Multiple Robots}
\label{subsection:learningcontinuouspols}
To overcome the aforementioned drawbacks in treating each robot as an independent entity, a small modification to the critic function (action-value) during training time is needed. During training, the critic function for robot $n$ uses some extra information $h$ from all other robots. This has been proposed in~\cite{maddpg,foerster2017counterfactual}. The modified action value function for robot n can then be represented as $Q_n((h_1(t),\ldots,h_N(t)), (a_1(t),\ldots,a_N(t)))$. The most naive method is to simply set $h_n(t) = \textbf{o}_n(t)$. 

Let policy for robot $n$ parameterized by $\theta_n$ be $\pi_{n}^{\theta_n}$.
For brevity sake, let $\{h_1(t),\ldots,h_N(t)\} = \mathbf{H}$, $\{a_1(t),\ldots,a_N(t)\} = \mathbf{A}$ and  ${\Pi} = \{\pi_{1}^{\theta_1},\ldots,\pi_{N}^{\theta_N}\}$
Thus in multi-robot case, the gradient of the actor function for robot $n$ is given as
\begin{equation}
\label{eq:madpggradient}
    \hat{g}_n = {\mathbb{E}}_t[\nabla_{\theta_n}\text{log} \pi_{n}^{\theta_n}(a_n(t)|\textbf{o}_n(t))\nabla_{a_n} \hat{Q}_n ^{\Pi}(\mathbf{H},\mathbf{A})]
\end{equation}
where $\hat{Q}_n ^{\Pi}(\mathbf{H},\mathbf{A})$ is the centralized critic function for robot $n$ that takes in input all robot observations and all robot actions and outputs a q value for robot $n$. The robot $n$ then takes a gradient of this q value with respect to to the action $a_n$ executed by robot $n$ and this gradient along with the policy gradient of robot $n$'s policy is used to update the actor function for robot $n$. It is important to note that the extra information from other robots actions is only used during training to update the critic function. This gives rise to centralized training but decentralized policies during inference. Thus, we now have a policy gradient algorithm that attempts to learn a policy for each robot such that \textbf{Problem 1} captured in Eqn. \ref{eq:rl} is maximized. 
\subsection{Backup Policies for Safety}
When using deep RL, one often loses any guarantees of safety. Thus, when attempting to maximize the reward in Eqn. \ref{eq:rl}, we have no guarantee that actions generated by our actor network are collision free (satisfy constraint in Eqn. \ref{eq:collision}). In real world applications of robotics this could be simply infeasible. Instead, we propose use of a simple analytical backup policy that ensures collision free trajectories. 

\begin{figure}[b!]
  \centering
  \includegraphics[scale = 0.35]{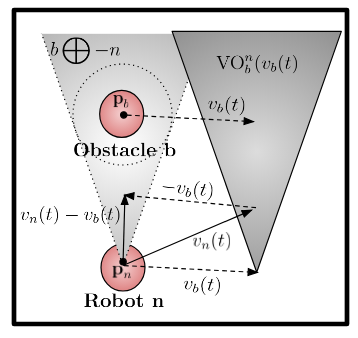}
  \caption{\textbf{Velocity Obstacle} Velocity obstacle $\text{VO}_b^n(v_b(t))$ of obstacle $b$ to robot $n$. When there exist multiple obstacles, the VO is defined as the union of all velocity obstacles.}
  \vspace{-0.2cm}
  \label{fig:vo}
\end{figure}

We use the Velocity Obstacle concept introduced in~\cite{fiorini1998motion}. While there exist more sophisticated algorithms for collision avoidance such as ORCA~\cite{van2011reciprocal} and NH-ORCA~\cite{alonso2013optimal}, we opt for VO due to its simplicity.
Consider a robot $n$, operating in the plane with its reference point at $\textbf{p}_n$ and let another planar  obstacle b (another robot or a static obstacle), be at $\textbf{p}_b$,  moving at velocity $v_b(t)$. The velocity obstacle $VO_b^n(v_b(t))$ of obstacle $b$ to robot $n$ is the set consisting of all those velocities $v_n(t)$ for robot $n$ that will result in a collision at some moment in time with obstacle $b$. The velocity obstacle (VO) is defined as: 
\begin{equation*}
\text{VO}_b^n(v_b(t)) = \{v_n(t) | \lambda(\textbf{p}_n, v_n(t)-v_b(t)) \cap b
\bigoplus - n \neq 0 \}    
\end{equation*}


where $\bigoplus$ gives the Minkowski sum between object $n$ and object $b$, $-n$ denotes
reflection of object $n$ reflected in its reference point $\textbf{p}_n$, and $\lambda(\textbf{p}_n, v_n(t)-v_b(t))$ represents a ray starting at $\textbf{p}_n$ and heading in the direction of the relative velocity of robot $n$ and b given by $v_n(t)-v_b(t)$.
~\cite{fiorini1998motion} show that the VO partitions the absolute velocities of robot $n$ into avoiding and colliding velocities. This implies that if $v_n(t) \in VO_b^n(v_b(t))$, then robot $n$ and obstacle $b$ will collide at some point in time. If $v_n(t)$ is chosen such that it is outside the VO of $b$, both objects will never collide and if $v_n$ is on the boundary, then it will brush obstacle b at some point in time.This concept is illustrated in Fig. \ref{fig:vo}. 


Each robots actor network outputs a linear force and a torque, i.e $a_n(t) = \{\textbf{F}_n(t),\tau_n(t)\}$. The dynamics of the robot then evolve according to :
\begin{equation}
\label{eq:dynamics}
    [v_n(t+1)] :=  \bigg[\frac{\textbf{F}_n(t) + z}{\kappa}\bigg](\Delta)
\end{equation}
\begin{equation}
\label{eq:posdynamics}
    [x_n(t+1),y_n(t+1)] :=  v_n(t+1)\Delta + [x_n(t),y_n(t)]
\end{equation}
where $z$ is some normally distributed noise $z \sim \mathcal{N}(0,\sigma^2)$, $\kappa$ is mass of the robot and is some fixed constant and $\Delta$ is the fixed time interval. Similarly, the rotational acceleration is derived from the torque and integrated over twice to update the orientation. For simplicity, say that the observation is set to just the position and the velocity in the 2D plane, i.e $\textbf{o}_t=[\textbf{p}_t,v_n(t)]$. From Eqn.~\ref{eq:dynamics}, we derive the  stationary  dynamics distribution with conditional density as $p(\textbf{o}_{t+1}|\textbf{o}_t,a_t) \sim \mathcal{N}(\frac{\Delta^2 \textbf{F}(t)+x_t}{\kappa},\frac{\sigma^2\Delta^4}{\kappa^2})$

A fundamental assumption for the existence of the gradient in  Eqn.~\ref{eq:ddpggradient} (and by extension, Eqn.~\ref{eq:madpggradient}) is that the conditional probability distribution $p(\textbf{o}_{t+1}=\xi|\textbf{o}_t,a_t)$ be continuous wrt $\textbf{o}_t$, $\textbf{a}_t$ $\forall \xi$ . We observe that in the case when the state evolves according to Eqn.~\ref{eq:dynamics}, the probability distribution is simply a gaussian distribution and is continuous. In order to incorporate the VO as a backup policy, we need to prove that the new transition function is still continuous. 
From Eqn.~\ref{eq:dynamics}, we have the velocity of the robot. Further, from Eqn. \ref{eq:posdynamics} $x_{t+1} = \Delta v_n(t)+x_t$. Since $x_t$ is a fixed, it suffices to find the continuity of $p(v_n(t))$ to conclude about the continuity of $p(\textbf{o}_{t+1}|\textbf{o}_t,a_t)$.
Consider an obstacle $B$ inside sensing range of robot $n$. To ensure safety, at every timestep, we compute the VO and check if the velocity $v_n(t) \in VO_B^n(v_b(t))$. In case, the velocity computed by the actor network falls inside the VO, we project the velocity back to the safe set $VO'$ ($VO'$ is the complement of the VO). The easiest projection can be given as :
\begin{equation}
    P^{min}_{{VO}'}(v_n(t)) = \{\min_{\Bar{v}} ||v_n(t)-\Bar{v}|| :\Bar{v}\in VO'\} 
    \label{eq:minproj}
\end{equation}
Thus, the safe velocity for robot $n$ is then given as:
\begin{equation}
\label{eq:safevel}
    v^{\text{safe}}_n(t)= v_n(t)\mathds{1}_{VO'}(v_n(t)) +  P^{min}_{{VO}'}(v_n(t))(1-\mathds{1}_{VO'}(v_n(t))
\end{equation}
where $\mathds{1}_{a}(b)$ is the indicator function and is $1$ if $b \in a$ and $0$ otherwise.
However, an issue with such a projection is that this gives us a discontinuous distribution for $p(v^{\text{safe}}_n(t))$ (probability) because now, there exists a set of values that $v^{\text{safe}}_n(t)$ never takes. Thus, this fails the assumption of smooth transition functions necessary to compute deterministic policy gradients. Additionally, in most real world systems, it is infeasible to make large changes to the velocity instantaneously.

\begin{figure}[t!]
  \centering
  \includegraphics[scale = 0.3]{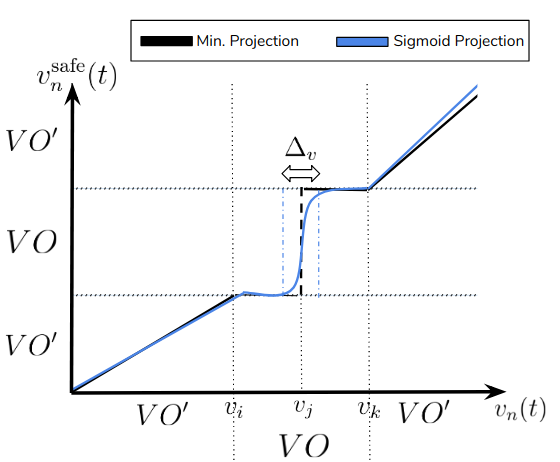}
  \caption{\textbf{Min Projection vs Sigmoid Projection of velocity} When using the minimum projection given in Eq ~\ref{eq:minproj} the safe velocity has a discontinuity. However, if we assume that the event $v_n(t) = v_j \pm \Delta_v $ occurs with very low probability, we get $p(v^{\text{safe}}_n(t))$ a continuous function.}
  \vspace{-0.2cm}
  \label{fig:smoothvsmins}
\end{figure}

To overcome this, we propose an alternate projection that ensures a smooth distribution of $v^{\text{safe}}_n(t)$. We note that at any given time $t$, the RVO set, i.e the set of infeasible velocities is always a continuous set. By exploiting this property, we propose the following alternative projection:
\begin{equation}
\label{eq:sigmoidprojection}
    P^{sig}_{{VO}'}(v_n(t)) = \frac{v_i -v_k}{1+e^{-c(v_n(t)-v_j)}} + v_k 
\end{equation}
where $v_i,v_j,v_k$ are the first, middle and last elements of the VO respectively and c is a hyperparameter that depends on the how quickly the robots can change their velocities. This is a shifted sigmoid projection and is visualized in Fig. ~\ref{fig:smoothvsmins}. 
Using the sigmoid projection, the safe velocity for robot $n$ is given as :
\begin{equation}
\label{eq:safevelfinal}
    v^{\text{safe}}_n(t)= v_n(t)\mathds{1}_{VO'}(v_n(t)) +  P^{sig}_{{VO}'}(v_n(t))(1-\mathds{1}_{VO'}(v_n(t))
\end{equation}
and from this we conclude that $p(\textbf{o}_{t+1}|\textbf{o}_t,a_t)$ is a continuous probability distribution thus enabling us to take the gradients specified in Eqn. ~\ref{eq:madpggradient}.
We put all these parts into our system for learning unlabeled multi-robot planning with motion constraints and present the full algorithm in Algorithm 1 and in the rest of the paper we reference it as \textbf{MARL+RVO}

\begin{algorithm}
  \caption{\textbf{Learning Safe Unlabeled Multi-Robot Motion Planning (MARL+RVO)}}\label{algo}
  \begin{algorithmic}[1]
  	\Require Initial random policy network and critic networks for all robots $\Pi$ Replay buffer $D$,
      \For {episode = 1 to $C$ ($C>>1$)}
        \State construct every robot's initial state $\textbf{o}_t^n$ (Eq ~\ref{eq:state}).
        \For {t= 1 to max episode length}
        \State for each robot, compute $a_t^n = \pi_n(\textbf{o}_t^n) $  
        \State for each robot, guarantee \textit{safe} $a_t^n$ (Eqn.~\ref{eq:safevelfinal}) 
        \State for each robot, compute $\textbf{o}_{t+1}^n$ and reward $r(t)$.
        \State Store $\textbf{o}_{t+1}^n, a_t^n, \textbf{o}_t^n, r(t)$ in $D$
        \For {robot = 1 to N}
            \State Sample minibatch of samples from $D$
            \State Compute bellman error using  Eqn.~\ref{eq:bellman}
            \State Update critic network using bellman error. 
            \State Compute policy gradient $\hat{g}_n$ from Eqn. ~\ref{eq:madpggradient}
            \State Update actor network using SGD with $\hat{g}_n$
        \EndFor
        \EndFor
        \EndFor
  \end{algorithmic}
\end{algorithm}
\begin{figure*}[t]
  \includegraphics[width=\textwidth]{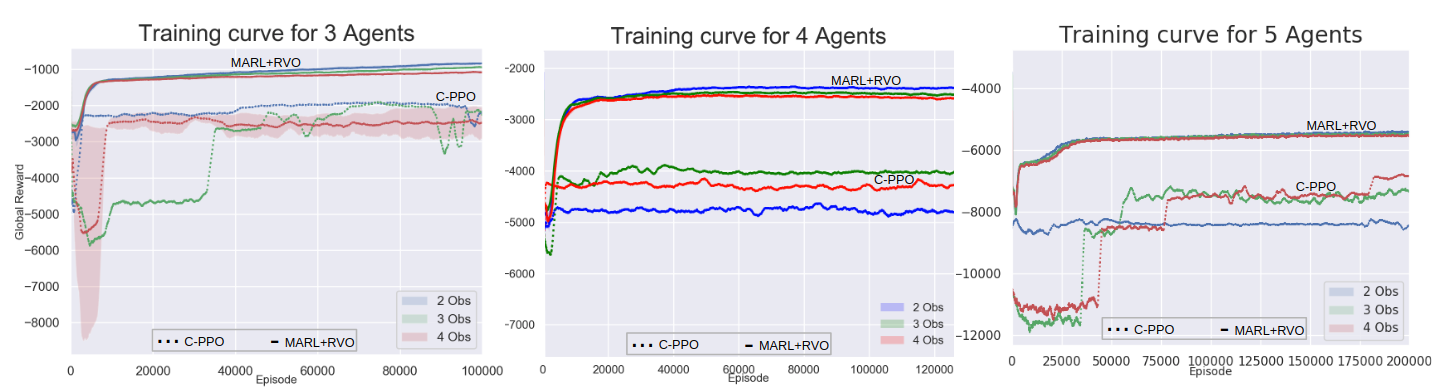}
  \caption{\textbf{Training curves for holonomic robots (with 2,3 and 4 Obstacles (Obs)).} We observe that the proposed MARL+RVO algorithm is able to converge and perform better than a centralized RL (C-PPO) policy. The global reward scale is different for each plot since it is a function of the space the robots operate in. Each curve is produced by running three independent runs of the algorithm. Darker line represents mean and shaded area represents
mean $\pm$ standard deviation of mean. }\label{fig:trainingcurveholonomic}
\vspace{-0.3cm}
\end{figure*}
\vspace{-0.2cm}
\section{Experimental Results}\label{sec:exp}
The efficacy of our algorithm is tested in simulated robotics experiments. We experiment by changing the number of robots, number of obstacles present in the environment and the robot dynamics.  In order to choose a meaningful reward function that ensures all goals are covered  we first compute for each goal the distance to its nearest robot. Then among this set of distances, we pick the maximum, negate it and add it to the reward. This represents the part of the reward function that forces all robots to cover all goals (denoted by $r_D(t)$). A similar strategy is adopted to ensure that the cosine difference between orientations is minimized (denoted by $r_r(t)$). In order to not overly depend on the projected velocity, we add in a negative penalty every time the projection to the safe set needs to be computed. Thus, we add a negative reward to all robots (denoted by $r_C(t)$). Thus, the overall reward given to each robot at time t is : 
\begin{equation}
    r(t) = \lambda_D r_D(t) + \lambda_r r_r(t) + \lambda_C r_C(t)
\end{equation}
where $ \lambda_D, \lambda_r$ and $\lambda_C$ are coefficients to balance each part of the reward function. This global reward function is the same for every robot operating in the environment. Maximizing this global reward requires a collective effort from all robots. 

It is important to note that during, inference time to guarantee safety, we do away with the soft projection introduced in Eqn. ~\ref{eq:sigmoidprojection}  and instead use the min projection as given in Eqn.~\ref{eq:minproj}. This is because during inference we no longer need to take gradients and hence the transition function need not be smooth continuous anymore. 
\begin{figure}[htb!]
  \centering
  \includegraphics[scale = 0.3]{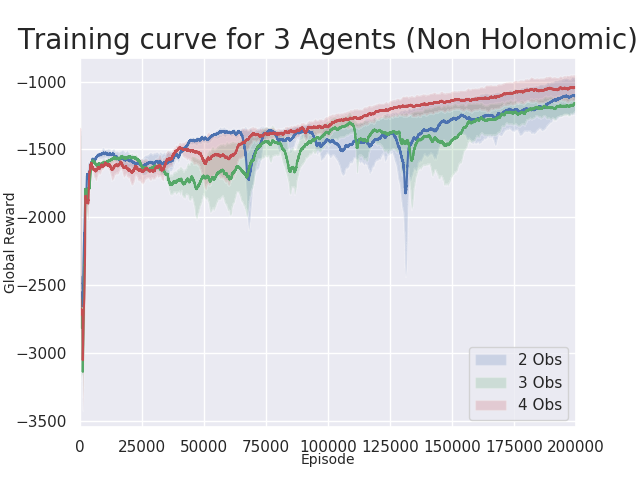}
  \caption{\textbf{Training curves for non holonomic robots (with 2,3 and 4 Obstacles (Obs)} When the robot dynamics are changed, MARL+RVO is still able to converge without making any changes to the loss function or the training parameters. }
  \vspace{-0.3cm}
  \label{fig:trainingcurvenonholo}
\end{figure}
\subsection{Experimental Setup Details}
For each robot, we setup an actor and critic network. The actor network consists of a two layer fully connected multi-layer perceptron (MLP). The critic network is also based on a similar fully connected MLP. The number of units in the hidden layers are varied depending on the size of the problem being solved (additional units and hidden layers when number of robots or obstacles are increased). For each episode, we set a maximum episode length of 300 steps. To update our networks, we use Adam and the learning rate is varied depending on the experiment under consideration. The discount factor ($\gamma$) is set to $0.95$ We also make use of a replay buffer to make sure dependencies between samples are modelled. The size of the replay buffer is $10^5$ and the size of the minibatch sampled is $1024$. The actions from the neural networks represent accelarations for the robots.

\begin{figure*}[hbt!]
  \includegraphics[scale=0.44]{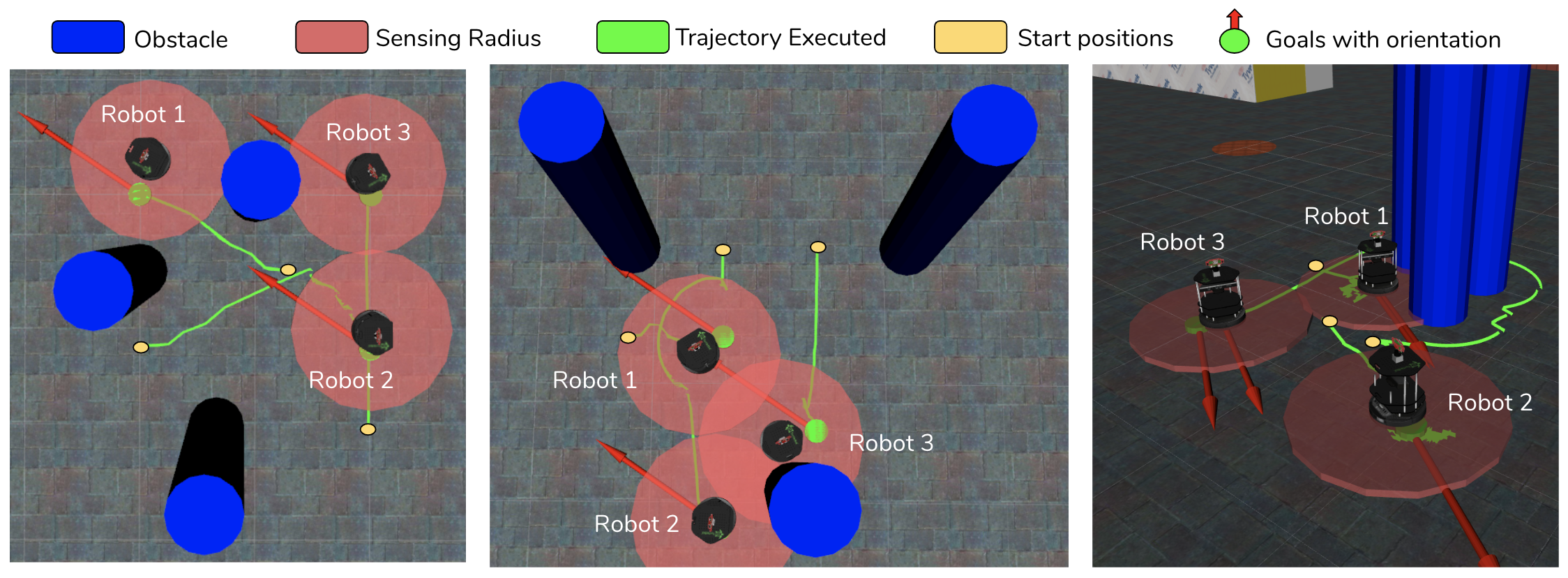}
  \caption{\textbf{Trajectory executed by 3 robots} Trajectories executed by robots in three randomly generated episodes after training is complete. In addition to robots reaching their goals in collision free manners, the proposed approach also aligns robots to desired final goal orientations.}\label{fig:trajplots}
 \vspace{-0.3cm}
\end{figure*}
The robots operate in a two dimensional space and do not have access to the third dimension. The space under consideration stretches from -1 unit to 1 unit in both the X and Y direction. Each robot is considered homogeneous and has nonzero mass. The radius of the robot is set to $0.05$ units. At the start of every episode obstacles, goals and start positions of robots are randomly populated. Radius of the obstacles is 0.12 units and the goal regions have a radius of 0.02 units. Robots are equipped with a sensor that returns perfect information (no noise in sensor measurements) about the pose and velocity of entities within the sensing range which is set at radius of 0.2 units. While the learning algorithm does not have an explicit assignment of goals in the states or in the reward function, when using RVO to avoid collisions, we need to greedily assign goals to agents based on distance to nearest goal and break all ties by randomly choosing goals. Once the RVO subprocess is done running, we again have no notion of goal assignment. In our simulated experiments (below) we set all units to meters.

\begin{figure*}[b]
  \vspace{-0.3cm}
  \includegraphics[width=\textwidth]{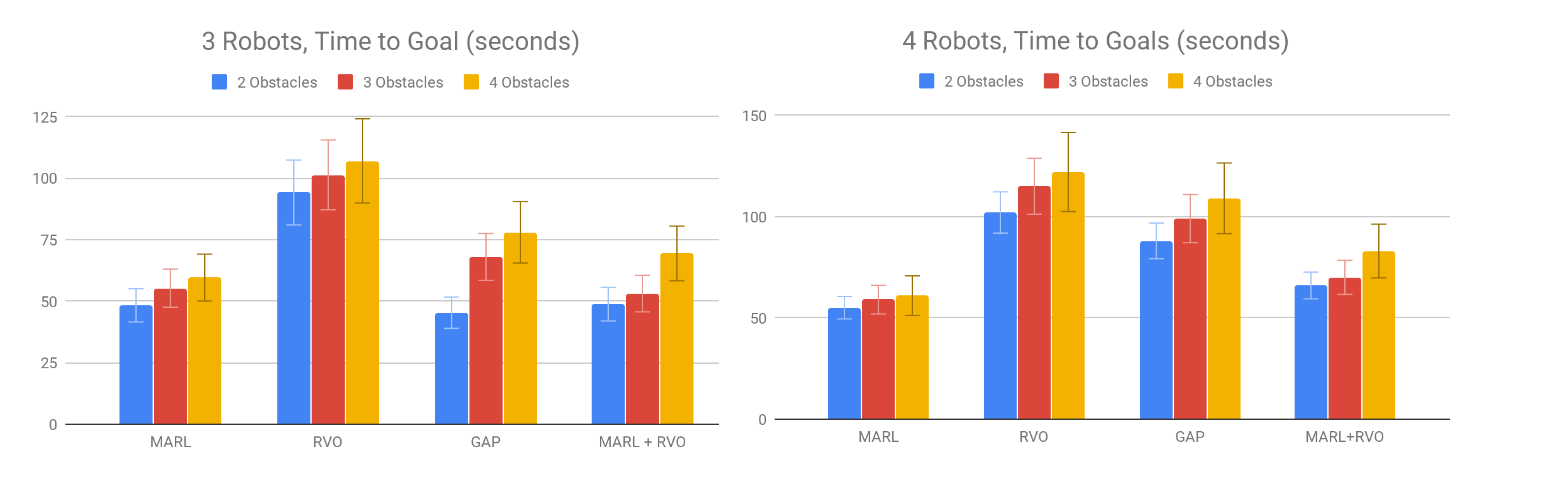}
  \caption{\textbf{Time taken to reach goal (with 2,3 and 4 Obstacles (Obs)) over 500 runs.} We compare with Multi-Agent Reinforcement Learning (MARL), Reciprocal Velocity Obstacles (RVO), GAP ~\cite{turpin2014goal} and our algorithm which combines the RL and safety(MARL+RVO). For the RVO method, we assign each robot its nearest goal (in terms of euclidean distance). GAP uses discrete nodes to search through space and hence its performance is contingent on the discretization of our continuous space. We observe that with MARL gives the best time performance, but this performance is not guaranteed to be collision free. Our method (MARL+RVO) trades-off time performance for guaranteed collision free trajectories.  }\label{fig:timeplots}
 \vspace{-0.3cm}
\end{figure*}

\subsection{Simulation Results}
We first observe from Fig. \ref{fig:trainingcurveholonomic} that the proposed MARL+RVO algorithm is able to converge even when the number of robots and the number of obstacles are increased. One of the key strengths of using a learning based solution for concurrent goal assignment and planning is that the algorithm can be used even when the dynamics of the robot change. When robot dynamics are changed to that of a non holonomic robot, we observe that our algorithm still converges.
This can be seen in Fig. \ref{fig:trainingcurvenonholo}. A simulated experiment setup is shown in Fig \ref{fig:trajplots}. In Fig. \ref{fig:trajplots} (left), a simple instance is shown where all robots must execute mostly straight line trajectories to arrive at goals. In Fig. \ref{fig:trajplots} (center), an interesting interaction takes place between Robots 1 and 2. Robot 2 takes a longer path curved path around Robot 1. Lastly, in Fig. \ref{fig:trajplots} (right) Robot 3 chooses to take a longer path around the obstacles in order to not cutoff Robot 2's path. These locally sub optimal, globally optimal behaviors are induced by using a global reward. We also design a centralized RL controller that uses information from all the agents and outputs a distribution for each agent. This controller is trained using PPO \cite{schulman2017proximal} and also uses velocity obstacles as a backup policy. We call this method Centralized PPO (C-PPO). We observe that C-PPO is unable to converge to an acceptable goal coverage policy.

In the RL framework, the policy attempts to maximize the reward function in a fixed horizon of time $T$. Thus, inherently the policy is being optimized for minimum time. To demonstrate this we compare our algorithm with vanilla MADDPG or MARL as described in \cite{maddpg}, reciprocal velocity obstacles \cite{van2008reciprocal} (RVO), and Goal Assignment and Planning (GAP) as introduced in \cite{turpin2014goal}.  
The RVO framework improves over VO. However, it is not a full "goal assignment and planning" framework and only generates collision free trajectories once goals have been assigned to robots. To benchmark, we assign goals in a greedy fashion. Each robot is assigned the goal closest to it. GAP utilizes a similar assignment but needs a discretization of the state space and a priority sequence for robots/goals. This prioritization is assigned randomly and the space is discretized into units of 0.1m. Out of these three methods, only RVO, GAP and MARL+RVO are guaranteed to produced collision free trajectories. For a fair comparison in terms of  time, we only consider those runs from MARL where no collision occurred. Our results are shown in Fig. \ref{fig:timeplots}. It can be seen that MARL and MARL+RVO is faster or almost comparable to GAP and RVO without needing any of the requirements of (assigning goals/discretized state space/priority sequence) GAP and RVO. Vanilla MARL is faster than MARL+RVO but isn't guaranteed to generate safe trajectories as seen from Table \ref{table:cap1}. 

\begin{table}[htb]
\begin{tabular}{|l|c|c|c|}
\hline
                  & \multicolumn{1}{l|}{3 Robots} & \multicolumn{1}{l|}{4 Robots} & \multicolumn{1}{l|}{5 Robots} \\ \hline
MARL              & 84                            & 192                           & 354                            \\ \hline
\textbf{MARL+RVO} & \textbf{0}                    & \textbf{0}                    & \textbf{0}                    \\ \hline
\end{tabular}
\caption{\textbf{Number of collisions for 3 robots in presence of 3 obstacles over 500 runs.}}\label{table:cap1}
\vspace{-0.1cm}
\end{table}

\begin{figure*}[t!]
  \centering
  \includegraphics[width =\textwidth]{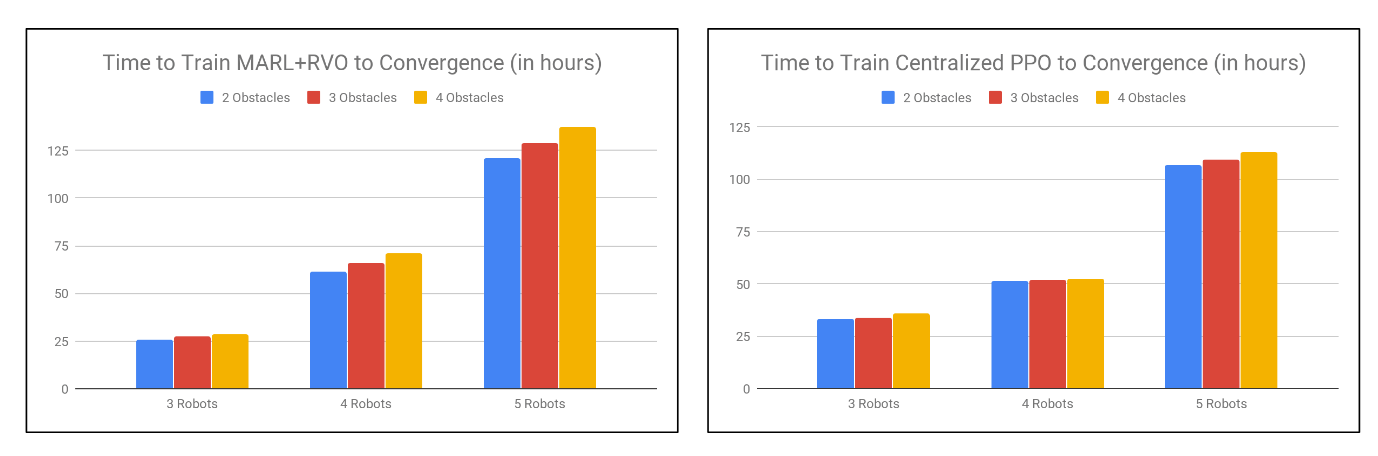}
  \caption{\textbf{Training time to convergence.} Training time for different configurations of robots and agents when trained on a NVIDIA DGX-1 (Tesla V100, 32GB $\times$ 8)}
  \vspace{-0.3cm}
  \label{fig:convergencetime}
\end{figure*}
\vspace{-0.3cm}
\section{Discussion}
In this paper we propose to solve the concurrent goal assignment and planning problem using MARL instead. Traditional approaches to solve this problem utilize a carefully designed heuristic function which produces guaranteed safe trajectories but breaks down if any of the assumptions are not satisfied. These assumptions restrict the class of problems that can be solved by traditional algorithms. By utilizing RL, we remove any assumptions on the robot dynamics or assumptions on the environment and instead use a global reward function that forces robots to collaborate with each other in order to maximize the reward. To overcome the lack of any safety guarantees, we propose using a model based policy in conjunction with the RL policy thus ensuring safe collision free trajectories. We demonstrate the effectiveness of our algorithm on simulations with varying number of obstacles, varying number of robots and varying robot dynamics and show that our proposed algorithm works faster and more robustly than traditional algorithms. 
\subsection{Caveats}
While this work attempts to learn an approximate solution for the unlabeled multi-robot problem, it has a few caveats. One of the biggest drawbacks of our work is that there is a significant engineering effort required in scaling up the number of robots. When the number of robots is increased, there are two major challenges that hamper MARL. The first is that the input space of the critic function grows as the number of robots increase. This increase in dimensionality necessitates longer training times for the critic. It might be possible to instead propose a local critic function that only takes in information from nearby robots. This might be possible by thinking of the robots as nodes on a graph and leveraging advances in graph neural networks \cite{kipf2016semi}, instead of using a fully connected network. 
The second problem is concerned with the need for more exploration as the number of robots increase. We observe from Fig \ref{fig:convergencetime} that the time required to train the algorithm grows almost exponentially as the number of robots are increased. While there exist massively parallel methods \cite{horgan2018distributed,espeholt2018impala} and software libraries \cite{liang2018rllib} to scale up for reinforcement learning, scaling up the number of robots still poses a significant computing challenge. Methods attempting to learn hierarchical policies for agents such as those in~\cite{khan2018scalable} might instead prove to be a suitable alternative. 
Lastly, our choice of VO for collision avoidance while rooted in its simplicity suffers from drawbacks many of which have been improved over by methods presented in \cite{van2011reciprocal}, \cite{alonso2018cooperative}. Adapting the projection step with more sophisticated collision avoidance algorithms is something we intend to explore in future work.

\addtolength{\textheight}{-12cm}   




\bibliographystyle{IEEEtran}
\bibliography{root}

\begin{thebibliography}{10}
\providecommand{\url}[1]{#1}
\csname url@samestyle\endcsname
\providecommand{\newblock}{\relax}
\providecommand{\bibinfo}[2]{#2}
\providecommand{\BIBentrySTDinterwordspacing}{\spaceskip=0pt\relax}
\providecommand{\BIBentryALTinterwordstretchfactor}{4}
\providecommand{\BIBentryALTinterwordspacing}{\spaceskip=\fontdimen2\font plus
\BIBentryALTinterwordstretchfactor\fontdimen3\font minus
  \fontdimen4\font\relax}
\providecommand{\BIBforeignlanguage}[2]{{%
\expandafter\ifx\csname l@#1\endcsname\relax
\typeout{** WARNING: IEEEtran.bst: No hyphenation pattern has been}%
\typeout{** loaded for the language `#1'. Using the pattern for}%
\typeout{** the default language instead.}%
\else
\language=\csname l@#1\endcsname
\fi
#2}}
\providecommand{\BIBdecl}{\relax}
\BIBdecl

\bibitem{desai2001modeling}
J.~P. Desai, J.~P. Ostrowski, and V.~Kumar, ``Modeling and control of
  formations of nonholonomic mobile robots,'' \emph{IEEE transactions on
  Robotics and Automation}, vol.~17, no.~6, pp. 905--908, 2001.

\bibitem{alonso2015multi}
J.~Alonso-Mora, S.~Baker, and D.~Rus, ``Multi-robot navigation in formation via
  sequential convex programming,'' in \emph{Intelligent Robots and Systems
  (IROS), 2015 IEEE/RSJ International Conference on}.\hskip 1em plus 0.5em
  minus 0.4em\relax IEEE, 2015, pp. 4634--4641.

\bibitem{saldana2016dynamic}
D.~Saldana, R.~J. Alitappeh, L.~C. Pimenta, R.~Assun{\c{c}}ao, and M.~F.
  Campos, ``Dynamic perimeter surveillance with a team of robots,'' in
  \emph{Robotics and Automation (ICRA), 2016 IEEE International Conference
  on}.\hskip 1em plus 0.5em minus 0.4em\relax IEEE, 2016, pp. 5289--5294.

\bibitem{adler2015efficient}
A.~Adler, M.~De~Berg, D.~Halperin, and K.~Solovey, ``Efficient multi-robot
  motion planning for unlabeled discs in simple polygons,'' in
  \emph{Algorithmic Foundations of Robotics XI}.\hskip 1em plus 0.5em minus
  0.4em\relax Springer, 2015, pp. 1--17.

\bibitem{macalpine2015scram}
P.~MacAlpine, E.~Price, and P.~Stone, ``Scram: Scalable collision-avoiding role
  assignment with minimal-makespan for formational positioning,'' in
  \emph{Twenty-Ninth AAAI Conference on Artificial Intelligence}, 2015.

\bibitem{yu2012distance}
J.~Yu and M.~LaValle, ``Distance optimal formation control on graphs with a
  tight convergence time guarantee,'' in \emph{Decision and Control (CDC), 2012
  IEEE 51st Annual Conference on}.\hskip 1em plus 0.5em minus 0.4em\relax IEEE,
  2012, pp. 4023--4028.

\bibitem{turpin2014capt}
M.~Turpin, N.~Michael, and V.~Kumar, ``Capt: Concurrent assignment and planning
  of trajectories for multiple robots,'' \emph{The International Journal of
  Robotics Research}, vol.~33, no.~1, pp. 98--112, 2014.

\bibitem{yu2013multi}
J.~Yu and S.~M. LaValle, ``Multi-agent path planning and network flow,'' in
  \emph{Algorithmic foundations of robotics X}.\hskip 1em plus 0.5em minus
  0.4em\relax Springer, 2013, pp. 157--173.

\bibitem{turpin2014goal}
M.~Turpin, K.~Mohta, N.~Michael, and V.~Kumar, ``Goal assignment and trajectory
  planning for large teams of interchangeable robots,'' \emph{Autonomous
  Robots}, vol.~37, no.~4, pp. 401--415, 2014.

\bibitem{maddpg}
R.~Lowe, Y.~Wu, A.~Tamar, J.~Harb, O.~P. Abbeel, and I.~Mordatch, ``Multi-agent
  actor-critic for mixed cooperative-competitive environments,'' in
  \emph{Advances in Neural Information Processing Systems}, 2017, pp.
  6379--6390.

\bibitem{foerster2017counterfactual}
J.~Foerster, G.~Farquhar, T.~Afouras, N.~Nardelli, and S.~Whiteson,
  ``Counterfactual multi-agent policy gradients,'' \emph{arXiv preprint
  arXiv:1705.08926}, 2017.

\bibitem{littman1994markov}
M.~L. Littman, ``Markov games as a framework for multi-agent reinforcement
  learning,'' in \emph{Machine Learning Proceedings 1994}.\hskip 1em plus 0.5em
  minus 0.4em\relax Elsevier, 1994, pp. 157--163.

\bibitem{lillicrap2015continuous}
T.~P. Lillicrap, J.~J. Hunt, A.~Pritzel, N.~Heess, T.~Erez, Y.~Tassa,
  D.~Silver, and D.~Wierstra, ``Continuous control with deep reinforcement
  learning,'' \emph{arXiv preprint arXiv:1509.02971}, 2015.

\bibitem{sutton1998reinforcement}
R.~S. Sutton and A.~G. Barto, \emph{Reinforcement learning: An
  introduction}.\hskip 1em plus 0.5em minus 0.4em\relax MIT press Cambridge,
  1998, vol.~1, no.~1.

\bibitem{dqn}
V.~Mnih, K.~Kavukcuoglu, D.~Silver, A.~Graves, I.~Antonoglou, D.~Wierstra, and
  M.~Riedmiller, ``Playing atari with deep reinforcement learning,''
  \emph{arXiv preprint arXiv:1312.5602}, 2013.

\bibitem{tan1993multi}
M.~Tan, ``Multi-agent reinforcement learning: Independent vs. cooperative
  agents.''

\bibitem{fiorini1998motion}
P.~Fiorini and Z.~Shiller, ``Motion planning in dynamic environments using
  velocity obstacles,'' \emph{The International Journal of Robotics Research},
  vol.~17, no.~7, pp. 760--772, 1998.

\bibitem{van2011reciprocal}
J.~Van Den~Berg, S.~J. Guy, M.~Lin, and D.~Manocha, ``Reciprocal n-body
  collision avoidance,'' in \emph{Robotics research}.\hskip 1em plus 0.5em
  minus 0.4em\relax Springer, 2011, pp. 3--19.

\bibitem{alonso2013optimal}
J.~Alonso-Mora, A.~Breitenmoser, M.~Rufli, P.~Beardsley, and R.~Siegwart,
  ``Optimal reciprocal collision avoidance for multiple non-holonomic robots,''
  in \emph{Distributed Autonomous Robotic Systems}.\hskip 1em plus 0.5em minus
  0.4em\relax Springer, 2013, pp. 203--216.

\bibitem{schulman2017proximal}
J.~Schulman, F.~Wolski, P.~Dhariwal, A.~Radford, and O.~Klimov, ``Proximal
  policy optimization algorithms,'' \emph{arXiv preprint arXiv:1707.06347},
  2017.

\bibitem{van2008reciprocal}
J.~Van~den Berg, M.~Lin, and D.~Manocha, ``Reciprocal velocity obstacles for
  real-time multi-agent navigation,'' in \emph{2008 IEEE International
  Conference on Robotics and Automation}.\hskip 1em plus 0.5em minus
  0.4em\relax IEEE, 2008, pp. 1928--1935.

\bibitem{kipf2016semi}
T.~N. Kipf and M.~Welling, ``Semi-supervised classification with graph
  convolutional networks,'' \emph{arXiv preprint arXiv:1609.02907}, 2016.

\bibitem{horgan2018distributed}
D.~Horgan, J.~Quan, D.~Budden, G.~Barth-Maron, M.~Hessel, H.~Van~Hasselt, and
  D.~Silver, ``Distributed prioritized experience replay,'' \emph{arXiv
  preprint arXiv:1803.00933}, 2018.

\bibitem{espeholt2018impala}
L.~Espeholt, H.~Soyer, R.~Munos, K.~Simonyan, V.~Mnih, T.~Ward, Y.~Doron,
  V.~Firoiu, T.~Harley, I.~Dunning \emph{et~al.}, ``Impala: Scalable
  distributed deep-rl with importance weighted actor-learner architectures,''
  \emph{arXiv preprint arXiv:1802.01561}, 2018.

\bibitem{liang2018rllib}
E.~Liang, R.~Liaw, R.~Nishihara, P.~Moritz, R.~Fox, K.~Goldberg, J.~E.
  Gonzalez, M.~I. Jordan, and I.~Stoica, ``{RLlib}: Abstractions for
  distributed reinforcement learning,'' in \emph{International Conference on
  Machine Learning ({ICML})}, 2018.

\bibitem{khan2018scalable}
A.~Khan, C.~Zhang, D.~D. Lee, V.~Kumar, and A.~Ribeiro, ``Scalable centralized
  deep multi-agent reinforcement learning via policy gradients,'' \emph{arXiv
  preprint arXiv:1805.08776}, 2018.

\bibitem{alonso2018cooperative}
J.~Alonso-Mora, P.~Beardsley, and R.~Siegwart, ``Cooperative collision
  avoidance for nonholonomic robots,'' \emph{IEEE Transactions on Robotics},
  vol.~34, no.~2, pp. 404--420, 2018.

\end{thebibliography}

\end{document}